# Learning Hyperspectral Feature Extraction and Classification with ResNeXt Network

Divinah Nyasaka[1], Jing Wang[1*], Haron Tinega[2]

[1]College of Software, Nankai University,  Tianjin, P.R.China
[2]School of Information Engineering, Zhengzhou University, Zhengzhou, P.R.China
*Corresponding author: Jing Wang  (e-mail: jingwang@nankai.edu.cn).

*Abstract*— **The Hyperspectral image (HSI) classification is a standard remote sensing task, in which each image pixel is given a label indicating the physical land-cover on the earth's surface. The achievements of image semantic segmentation and deep learning approaches on ordinary images have accelerated the research on hyperspectral image classification. Moreover, the utilization of both the spectral and spatial cues in hyperspectral images has shown improved classification accuracy in hyperspectral image classification. The use of only 3D Convolutional Neural Networks (3D-CNN) to extract both spatial and spectral cues from Hyperspectral images results in an explosion of parameters hence high computational cost. We propose network architecture called the MixedSN that utilizes the 3D convolutions to modeling spectral-spatial information in the early layers of the architecture and the 2D convolutions at the top layers which majorly deal with semantic abstraction. We constrain our architecture to ResNeXt block because of their performance and simplicity. Our model drastically reduced the number of parameters and achieved comparable classification performance with state-of-the-art methods on Indian Pine (IP) scene dataset, Pavia University scene (PU) dataset, Salinas (SA) Scene dataset, and Botswana (BW) datasets..**

## I. Introduction

Hyperspectral imaging is a remote sensing technique that involves the collection of electromagnetic spectrum reflected by the objects from the homogenous area on the earth's surface [1]. The electromagnetic spectrum bands ranging from visible to near-infrared wavelength are collected using the hyperspectral imaging sensor often mounted on aircraft or satellites [2]. The development of hyperspectral imaging sensors has resulted in the collection of voluminous spectral information derived from hundreds of bands, hence the development of spectral-based feature extraction systems [3].

The availability of detailed spectral information coupled with the high spatial correlation between different channels from the same area conveys useful information that is vital in HSI image classification [1] [4] [5] [6]. However, the multiple bands in Hyperspectral images suffer from the curse of dimensionality as they contain voluminous data resulting in increased dimensionality of the images in the spectral domain. Therefore, conventional techniques developed for hyperspectral image analysis are rendered inefficient [7]. To address the curse of dimensionality, feature extraction (FE) is regarded as an important phase in Hyperspectral image processing [8]. Traditionally, a hyperspectral image feature extraction uses hand-designed techniques [1]. Due to the spatial variability of spectral signatures, the extraction of the most discriminative features or bands is still a challenging task [9].

Inspired by the ability of deep learning to extract discriminative features without much preprocessing, many researchers began to study the use of deep learning in hyperspectral image feature extraction. This has improved the performance of hyperspectral image feature extraction systems [10]. The family of Inception models has demonstrated that network topology affects both model complexity and accuracy [11]. However, all Inception models have one common property which is the split-transform-merge strategy. Research work by Zhong et al [12] has also verified the importance of the residual network in HSI classification. It is in this perspective that we propose an architecture network called the MixedSN that utilizes the 3D convolutions to modeling spectral-spatial information in the early layers of the architecture and the 2D convolutions at the top layers which majorly deals with semantic abstraction. Our contribution includes the development of a cost-effective 3D-2D implementation network design topology for HSI images classification which achieves comparable classification performance with the state-of-the-art methods on IP dataset, PU dataset, SA Scene dataset, and BW dataset.

The rest of this paper is organized such that Section 2 presents the related work, Section 3 presents our proposed model, Section 4 presents the experiments while Section 5 contains the conclusion.

## II. Related Works

Early works of Hyperspectral image feature extraction (HSI FE) methods apply linear transformations to extract discriminative features from the spectral dimension of HSI data. These methods include  principal component analysis (PCA) [1], independent component analysis (ICA) [2], linear discriminant analysis [3] [4] [5] and classifiers such as linear

SVMs and logistic regression (LR) [6]. However, hyperspectral data are naturally nonlinear. The nonlinearity of hyperspectral data is caused by: 1) the undesired light-scattering mechanisms of other land cover objects such as vegetation which may distort the spectral characteristics of the object of interest. 2) the different atmospheric scattering caused by particles in the atmosphere [7] [8].The nonlinearity of hyperspectral data renders the use of only linear transformation-based methods unsuitable for their analysis. In addition, these methods are single layer learning methods that downgrade the capacity of feature learning, thus degrading overall feature learning accuracy [6].

To solve the nonlinearity challenge in hyperspectral data, researchers turned to manifold learning [9] in hyperspectral image processing. Manifold learning seeks to find the inherent structure of data that is nonlinearly distributed, which is highly useful for hyperspectral image feature extraction (HSI FE) [10]. Though supervised manifold learning variants exist, the typical manifold learning problem is unsupervised. Other researchers used the kernel-based algorithms to address the nonlinearity data challenge in hyperspectral data. Kernel methods transform the original data into a higher dimensional Hilbert space providing a possibility of mapping a nonlinear problem to a linear one [11].

To solve the challenge of single-learning layer methods, several hand-designed feature extraction approaches for classification were developed over time and they include the Sparse Self-Representation [12], Multiscale Super pixels and Guided Filter [13], Joint Sparse Model and Discontinuity Preserving Relaxation [14], Fusing Correlation Coefficient and Sparse Representation [15], Boltzmann Entropy-Based Band Selection [16]. Other researchers borrowed from the visual system of humans that employs a sequence of different phases of processing (on the order of 10) for object recognition tasks [17], to include more layers to extract new features. The kernel SVMs was the first to use the two-layered method developed to extract new features [6].

Later, the Convolutional Neural Networks (CNNs) was developed to effectively extract information from the spatial domain through the use of local connections. In order to prevent network parameters explosion, weights were shared among different network layers [18]. Over the years, researchers have sought to include more layers into CNN architecture to improve accuracy. The inclusion of more than three layers to extract new features resulted in the development of deep learning-based methods designed to simulate the process from the retina to the cortex [19]. A deep neural network (DNN) has the ability to represent complicated data. Deep learning involves a class of models that try to learn features and tasks directly from original data through a series of hierarchical layers. The earlier layers extract simple structures such as texture and edges, whereas the later layers represent this information into more complicated features. The extraction of high-level features from the low-level features leads to the extraction of abstract and invariant features, which makes deep learning suitable for a wide range of tasks such as classification and target detection [20] [19] [21]. Some of the deep learning methods for HSI images include convolutional neural networks (CNNs), deep belief networks (DBNs), stacked autoencoders (SAEs), recurrent neural networks (RNNs), and generative adversarial networks (GANs).

2D-CNN has achieved tremendous results in visual data processing such as face detection [22], semantic segmentation [23], image classification [24], [25] colon cancer classification [26], object detection [27] [40] and depth estimation [28] [38] [39]. However, the use of 2D-CNN in hyperspectral image analysis results in missing channel relationship information. Thus, 2D-CNN alone lacks the ability to extract good discriminative features from the spectral dimension. Unlike the 2D-CNN, the 3D-CNN has the ability to simultaneously extract the spectral and spatial information from hyperspectral data whilst achieving better accuracy as compared to 2D-CNN. However, 3D-CNN is computationally expensive to be used alone in hyperspectral image analysis. Moreover, 3D-CNN network topology is simple and lacks feature aggregations preventing the model from deepening to allow extraction of deep features. This explains why 3D-CNN tends to perform poorly in the classification of pixels of different classes with similar textures over many spectral bands. It is clear that the 2D-CNN and 3D-CNN have their share of challenges that when used alone prevents them from achieving better accuracy on hyperspectral images. However, when combined, they overcome the challenges and are able to achieve better accuracy in hyperspectral image analysis. Zhong et al [29] proposed Spectral-spatial Residual Network (SSRN) model that implements 3D-CNN residual network using ResNet[30] as the backbone architecture. The model revealed the possibility of deepening the 3D-CNN network to enhance the extraction of deep features which result in high HSI classification accuracy. However, the summation method used to aggregate feature at each ResNet layer requires layer output feature maps to have consistent scale as the residual feature maps, hence each ResNet layer has its own weights which overally lead to an explosion of network parameters. Roy et al [31], proposed HybridSN model that combines the 3D-CNN and 2D-CNN network architectures. The model achieved the best state-of-the-art accuracies on almost all the publicly available HSI datasets. The HybridSN model demonstrates that a well designed 3D-CNN with 2D-CNN simple network structure can still give good accuracy on HSI classification. However, in spite of this good accuracy, the model contains high number of parameters compared to SSRN model while on the other hand the SSRN model has longer training time.

## III. PROPOSED MIXEDSN MODEL

Our model combines the 3D-CNN and 2D-CNN layers to extract the spectral-spatial information encoded in multiple contiguous HSI bands. The model input is 3D cubes of the hyperspectral data, whereas the output is the 1-D label probability distribution vector. The framework consists of 3D-CNN at the bottom of the network, which facilitates the joint learning of spectral-spatial feature representation from a stack of hyperspectral image bands and the 2D-CNN at the top of 3D-CNN to further learn deep spatial representation.

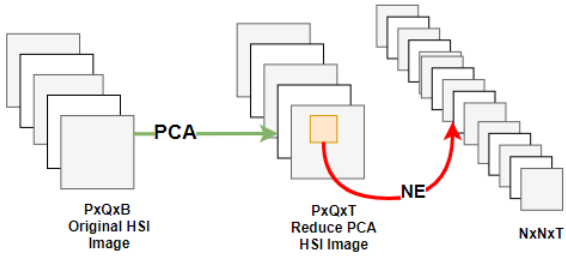

Fig. 1.Spectral-spatial Hyperspectral Input Preprocessing: N.E = Neighborhood Extraction. The Spectral Bands for IP, PU, SA, and BW datasets after PCA are 30,15,15,13 respectively.

The first step is the preprocessing of the original spectral-spatial hyperspectral input data cube $I \in R^{P \times Q \times B}$, where $P$, $Q$ and $B$ represents the width, height, and the depth (number of spectral bands), respectively. Therefore, each spectral-spatial hyperspectral image pixel in $I$ has the depth of $B$ thus forming a one-hot label vector $V = (v1, v2, \dots vL) \in R^{1 \times 1 \times L}$, where $L$ represents the land-cover categories. The input data cube $I$, therefore, contains high intra-class variability and interclass similarity as a result of mixed land–cover classes displayed by the spectral-spatial hyperspectral image pixel. To remove this spectral redundancy in $I$, we propose the application of the principal component analysis (PCA) over the raw input data cube $I$ along the spectral dimension. The PCA works by downsizing the raw input data cube's depth dimension from $B$ to $T$ while maintaining the width and the height dimensions, such that the reduced PCA spectral-spatial hyperspectral input data cube $P \in R^{P \times Q \times T}$. Here, $T$ is the number of spectral bands after PCA. For image classification purposes, the reduced PCA spectral-spatial hyperspectral input data cube $P$ is divided into n small overlapping 3D neighboring patches $Q \in R^{S \times S \times T}$ centered at the spatial location $(x, y)$. Here, $S \times S$ is the width and the height of the covering window while $T$ is the window depth. The label of the central pixel at the spatial location $(x, y)$ decides the truth labels. Therefore, the total number of generated 3D-patches (n) from S is given by $(P - S + 1) \times (Q - S + 1)$. Thus, $Q_{(x,y)}$ which is a 3D-patch at location $(x, y)$ covers the width from $x - (S - 1)/2$ to $x + (S - 1)/2$, the height from $y - (S - 1)/2$ to $y + (S - 1)/2$ and all $T$ spectral bands.

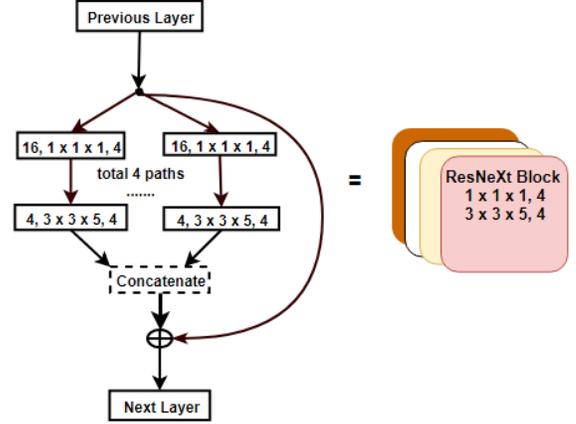

Fig. 2 ResNeXt Block that depicts the convolution used in the first 3D ResNeXt block ( see fig. 3). The block has a cardinality $C(C = 4)$. Cardinality is the total number of branching paths inside the ResNeXt block. Each layer is denoted by input channels, kernel size, and output channels. For example 16, 1 x 1 x 1, 4 represent 16 input channels, 1 x 1 x 1 filter size and 4 output channels. In this paper, we adopt 4 total paths which are denoted by 4 stacked rectangle with different colors (see the right fig). Also for simplicity, we omit the input channels inside each convolution, however, the total output channel are assumed to be matched with input channel as required by ResNeXt structure

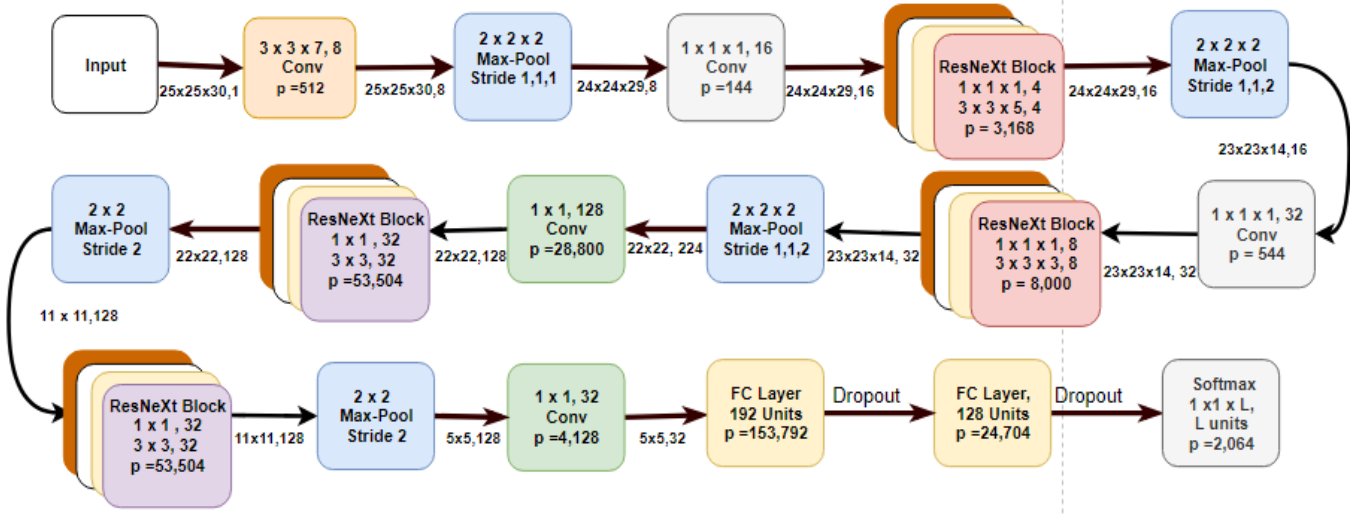

Fig. 3.The structure of our proposed MixedSN model: p = number of parameters , FC = Fully Connected Layer and L = 16 = number of classes . In addition, we apply zero padding at every convolution layer and each layer is denoted by filter size and output channels e.g 3 x 3 x7, 8 represents 3 x 3 x 7 filter and 8 output channels.

Our architecture implements the hypothesis that postulates that modeling spectral information (i.e., 3D convolutions) in the early layers (the ones closest to the pixels) might be vital as compared in the top layers. The top layers majorly deal with the semantic abstraction (i.e., 2D convolutions) where spectral modeling is not necessary. As a result, the lowest layers (the ones closest to the pixels) of our architecture network contains the 3D convolutions, while the top layers (the ones closest to the fully connected (FC) layers) contain the 2D convolutions (see Fig. 3). To capture the spectral-spatial information in the lower layers, a 3D kernel is stridden over a spectral-spatial hyperspectral input data cube such that the activation value at spectral-spatial position $(x, y, z)$ in the $j^{th}$ feature map of the $i^{th}$ layer denoted as $v_{i,j}^{x,y,z}$, is given by:

$$v_{i,j}^{x,y,z} = ReLU\left(b_{i,j} + \sum_{m=1}^{M} \sum_{p=0}^{P_i-1} \sum_{q=0}^{Q_i-1} \sum_{r=0}^{R_i-1} w_{i,j,m}^{p,q,t} \times v_{(i-1),m}^{(x+p),(y+q),(z+r)}\right) \quad (1)$$

Where parameters $P_i, Q_i, R_i$ is the width, the height, and the depth of the kernel, respectively. Here the depth of the kernel is the spectral dimension. Parameter $b_{i,j}$ is the bias value for the $j^{th}$ feature map of the $i^{th}$ layer, $M$ is the total number of feature maps in the $(i-1)^{th}$ layer connected to the current feature map. $w_{i,j,m}^{p,q,r}$ is the value of the weight parameter for position $(p, q, r)$ kernel connected to the $m^{th}$ feature map in the previous layer.

In the upper layers, the spatial feature learning part is done by convolving the input data from the previous layers with the 2D kernels resulting in 2D discriminative feature maps. To introduce nonlinearity in our model, the convolved feature maps are passed through the ReLU activation function such that the activation value at position $(x, y)$ in the $j^{th}$ spatial feature map of the $i^{th}$ CNN layer is symbolized as $v_{i,j}^{x,y}$ and can be generated using the equation:

$$v_{i,j}^{x,y} = R\left(b_{i,j} + \sum_{m=1}^{M} \sum_{p=0}^{P_i-1} \sum_{q=0}^{Q_i-1} w_{i,j,m}^{p,q} \times v_{(i-1),m}^{(x+p),(y+q)}\right) \quad (2)$$

Where $R$ is ReLU activation function, the value $w_{i,j,m}^{pq}$ is the weight parameter for spatial position $(p, q)$ kernel connected to the $m^{th}$ feature map in the previous layer.

We constrain our architecture to ResNeXt [32] block of cardinality 4 (see Fig. 2) because of their performance and simplicity. ResNeXt blocks are deep residual networks with cardinality. They use the split-transform-merge strategy that results in branching paths within a cell to transform the residual block. The output from ResNeXt block is added with the skip connection path. This results in an orthogonal increase in the depth of the residual networks. The number of branching paths inside the ResNeXt block is the cardinality of the block. Mathematically our ResNeXt block can be expressed as:

$$y = x + \sum_{i=1}^{C} \tau_i(x) \quad (3)$$

Where $x$ is the input from the previous layer, $y$ is the output, $C$ is the cardinality, and $\tau_i$ is the arbitrary conversion.

For our architecture, we have 3D convolution and pooling kernel of size $k \times k \times h$, and 2D convolution and pooling kernel of size $k \times k$, where $h$ is the depth of the kernel and $k$ is kernel's spatial size. In the lower layers (the ones closest to the pixels) of our architecture, we begin by convolving the raw input data cube using eight $3 \times 3 \times 7$ kernels followed by a max-pooling layer. Then, we used two 3D-ResNeXt blocks to achieve spectral-spatial feature learning. These two 3D-ResNeXt blocks are preceded by a 1x1x1 scale-up layer and succeeded by a 3D max-pooling. In the first and the second ResNeXt block, $\tau$ is composed of a continuous convolution (Conv $1\times1\times1$ → Conv $3\times3\times5$) and (Conv $1\times1\times1$ → Conv $3\times3\times3$) respectively. In the upper layers of the network just before the FCs layers, we have two 2D-ResNeXt blocks to further learn deep spatial encoded features and they are preceded by a $1 \times 1$ scale-down layer and succeeded by a 2D max-pooling. These two 2D-ResNeXt blocks contain the same parameters such that, $\tau$ is composed of a continuous convolution (Conv1 $\times1$→Conv3 $\times3$).

Each convolutional layer is applied with appropriate padding and stride 1 thus the input size is the same as the output. In addition, each convolutional layer is followed by a ReLU activation function to increase nonlinearity to address the problem of overfitting caused by limited training samples of hyperspectral data. ReLU also improves the capability of the model to represent complex functions and facilitates optimization resulting in lower training and testing losses. We used the max pooling to speed up the training process and achieve spatial invariance whilst maintaining accuracy. The basic idea of max pooling is to select the most discriminative feature and use it to represent a set of features. In the case of 2D pooling, the maximum value in the neighborhood $v_{i,m}^{x,y}$ is given by:

$$v_{i,m}^{x,y} = \max_{p \in [0, P_i-1], q \in [0, Q_i-1]} v_{(i-1),m}^{(x+p),(y+q)} \quad (5)$$

Where m indexes the feature map in the $(i-1)^{th}$ convolution layer and $P_i, Q_i$ is the kernel size. Whilst in the 3D- pooling the maximum value in the neighborhood $v_{i,m}^{x,y,z}$ is as follows;

$$v_{i,m}^{x,y,z} = \max_{p \in [0, P_i-1], q \in [0, Q_i-1], r \in [0, R_i-1]} v_{(i-1),m}^{(x+p),(y+q),(z+r)} \quad (6)$$

Where, $P_i, Q_i, R_i$ is the width, the height and the depth of the kernel.

In our experiment, we implement the max pooling with varying kernel sizes and strides. Specifically, there are five max pooling layers. The first pooling layer has a kernel size of $2 \times 2 \times 2$ and stride of 1. Since the spectral-spatial feature learning occurs in the lower layers and the deeper spatial feature learning in the upper layers of our architecture, the second and third 3D max pooling layers have the kernel size of $2 \times 2 \times 2$ and stride $1, 1, 2$. This is not to adversely interfere with the spatial features at the early stage of the network. The fourth and fifth 2D max pooling layers have a kernel size of 2 × 2 and stride of 2. We implemented the spatial bottlenecks at two points in our architecture i.e. when transiting from 3D to 2D and just before the FCs layers to drastically reduce the input feature maps and increase the training speed. Then the output is flattened before feeding to the FC layers that output the land cover class probabilities through the use of a softmax loss layer $l_o$ given by

$$l_o = -\frac{1}{p}\sum_{i=1}^{p}\sum_{j=1}^{j}[r_{ij}\log(q_{ij})] \quad (7)$$

Where, $j$ is the number of class labels, p denotes the mini-batch size, $q_i$ and, $r_i$ denotes the $i^{th}$ label probability distribution vector and the ground truth label in the minibatch, respectively. The average is done on the sum result from the whole mini-batch pixels.

The two fully connected layers have 192 and 128 outputs respectively with dropout layers. To address the problem of overfitting caused by insufficient training samples of HSI data and achieve better model generalizability, we use a 40% dropout rate on IP, PU and SA datasets, while on the BW dataset we apply dropout rate of 45% since it has very small sampled data.

## IV. EXPERIMENTATION

### A. Datasets

We evaluate our model's performance using four publicly available HSI datasets, which are the IP, PU, BW, and SA datasets. For each dataset, we randomly split the labeled samples into two subsets, i.e., training and test samples. Then we carried two tests for each dataset. In the first test, we randomly divide the dataset into 10% training and 90% testing. In the second test, we randomly divide the dataset into 30% training and 70% testing. Table 1 provides a summary description of each dataset used in this experiment.

TABLE 1:
SUMMARY OF HYPERSPECTRAL IMAGE DATASETS USED IN EXPERIMENTATION

| Dataset Name | Year | Source | SD (Pixels) | SB | WR(nm) | Labels | Classes | Mode | SR (m) |
|---|---|---|---|---|---|---|---|---|---|
| Indian Pines | 1992 | NASA AVIRIS | 145 x 145 | 220 | 400 - 2500 | 10249 | 16 | Aerial | 20 |
| Pavia University | 2001 | ROSIS-03 sensor | 610 x 610 | 115 | 430 - 860 | 42776 | 9 | Aerial | 1.3 |
| Salinas | 1998 | NASA AVIRIS | 512 x 217 | 224 | 360 - 2500 | 54129 | 16 | Aerial | 3.7 |
| Botswana | 2001-2004 | NASA EO-1 | 1496 × 256 | 242 | 400-2500 | 3248 | 14 | Satellite | 30 |

SD = Spectral Dimension, SB = Spectral Band, WR = Wave Length; SR = Spatial Resolution

**The Indian Pines dataset** was collected by Purdue University Research Repository (PURR) in 1992 using NASA's AVIRIS sensor flying over the Indian Pines test site in North West Indiana. The image has a spatial dimension of 145 x 145 pixels with 20 meters spatial resolution and 220 spectral bands in 400–2500 nm wavelength range. The samples in the Indian Pines scene image that contains no information together with 20 water absorption bands ([104-108], [150-163], 220) were discarded before adopting the image for analysis. The discarded samples are the unlabeled data presented as a black strip (see Fig. 5 (b, c, d)). The dataset's ground truth differentiates 16 classes (see table 2), which are not mutually exclusive.

Prof. Paolo Gamba of Pavia University, Italy collected the **Pavia University scene** dataset in 2001. The dataset consists of a hyperspectral image taken by the Reflective Optics System Imaging Spectrometer (ROSIS) sensor flying over the over Pavia city, northern Italy. The image has a spatial dimension of 610 x 340 pixels, 115 spectral bands in 430–860nm wavelength range, and 1.3 meters spatial resolution. Twelve (12) noisy bands and samples with no information were removed before the data was analyzed. The discarded samples are the unlabeled data presented as a black strip (see fig. 7 (b, c, d)). The Pavia University scene contains 9 classes, out of which the meadows class covers almost half of the entire dataset.

The **Salinas Scene dataset** was acquired by NASA's AVIRIS sensor flying over the Salinas Valley, California in October 1998. The image measures a spatial dimension of 512 x 217 pixels with 3.7 meters spatial resolution and 224 spectral bands from a wavelength range of 360–2500 nm. We reduced the number of bands from 224 to 214 by discarding [108-112], [154-167], 224) bands covering the water absorption region. We also discard samples in the Salinas Scene image that contains no information. The discarded samples are the unlabeled data presented as a black strip (see fig. 9 (b, c, d)). The Salinas Scene land-cover has been categorized into 16 class labels and the grape trees class covers the largest portion (a fifth) of the entire Salinas scene dataset.

Finally, the **Botswana dataset** was collected by The UT Center for Space Research from 2001 to 2004 using NASA EO-1 satellite flying over the Okavango Delta, Botswana. The data used in this experiment was acquired in June 2001. The dataset consists of 14 classes representing the equivalent number of land cover types in seasonal swamps, occasional swamps, and drier woodlands located in the distal portion of the Delta. The image has a spatial dimension of 1496 × 256 pixels with 30 meters spatial resolution and 242 spectral bands in 400–2500 nm wavelength range. Before employing the image for analysis, 97 uncalibrated and water corrupted bands were discarded resulting in a new depth dimension of 145 bands. In addition, the samples that contain no information were also removed and are presented as unlabeled data (see fig. 11 (b, c, d)).

TABLE 2:
NUMBER OF TRAINING AND TEST SAMPLES USED FOR THE INDIAN PINE P SCENE DATASET USING 10% TRAIN SAMPLE, 90% TEST SAMPLE, AND 30% TRAIN SAMPLE, 70% TEST SAMPLE

| Class No | Class Label | Samples (Pixels) | Sample Cover (%) | 10% train, 90% test | | 30% train, 70% test | |
|---|---|---|---|---|---|---|---|
| | | | | Train Sample | Test Sample | Train Sample | Test Sample |

| 1 | Alfalfa | 46 | 0.45 | 5 | 41 | 14 | 32 |
|---|---|---|---|---|---|---|---|
| 2 | Corn-Notill | 1428 | 13.93 | 143 | 1285 | 428 | 1000 |
| 3 | Corn-Mintill | 830 | 8.10 | 83 | 747 | 249 | 581 |
| 4 | Corn | 237 | 2.31 | 24 | 213 | 71 | 166 |
| 5 | Grass-Pasture | 483 | 4.71 | 48 | 435 | 145 | 338 |
| 6 | Grass-Trees | 730 | 7.12 | 73 | 657 | 219 | 511 |
| 7 | Grass-Pasture-Mowed | 28 | 0.27 | 3 | 25 | 8 | 20 |
| 8 | Hay-Windrowed | 478 | 4.66 | 48 | 430 | 143 | 335 |
| 9 | Oats | 20 | 0.20 | 2 | 18 | 6 | 14 |
| 10 | Soybean-Notill | 972 | 9.48 | 97 | 875 | 292 | 680 |
| 11 | Soybean-Mintill | 2455 | 23.95 | 245 | 2210 | 736 | 1719 |
| 12 | Soybean-Clean | 593 | 5.79 | 59 | 534 | 178 | 415 |
| 13 | Wheat | 205 | 2.00 | 20 | 185 | 62 | 143 |
| 14 | Woods | 1265 | 12.34 | 126 | 1139 | 379 | 886 |
| 15 | Buildings-Grass-Trees-Drives | 386 | 3.77 | 39 | 347 | 116 | 270 |
| 16 | Stone-Steel-Towers | 93 | 0.91 | 9 | 84 | 28 | 65 |

### B. Experimental Setup

Our experiments were run on Google Inc. online cloud service (Colab) platform with 25GB RAM and 1 GPU. After designing our network, we analyzed the various factors that affect the training process and model performance. These factors include input spatial window size, Dropout rate, learning rate, and the number of epochs. We employ a greedy search method to set our model optimal hyper-parameters. We set the learning rate at 0.001 with a weight decay rate of 1e-06, and a 40% dropout rate on Indian Pines, Pavia University and Salinas scene datasets. However, on the Botswana dataset, we slightly increase the dropout rate to 45% due to the very small sampled data. Then, the network weights were randomly initialized and trained by Adam gradient descent optimizer method with a softmax loss function. Each experiment was run for 100 epochs. For a fair comparison with other state-of-the-art methods, we adopt a spatial window size of $25 \times 25$ which is equivalent to the one used in the HybridSN model.

### C. Result and Analysis

The results are reported and analyzed at the dataset level. First, we present per-class accuracy, and then we use the Overall Accuracy (OA), Average Accuracy (AA) and Kappa Coefficient (Kappa) evaluation criteria to assess the overall performance of various approaches. Overall Accuracy (OA) represents the percentage of correctly classified samples, while the Average Accuracy (AA) gives the mean result of per-class classification accuracies. The Kappa Coefficient (Kappa) provides information on what percentage the classification map concur with the ground truth map. We compare all the datasets accuracies with the state of art-of-art methods such as SVM [33], 2D-CNN [34], 3D-CNN [35], M3D-CNN [36], SSRN [29], and HybridSN[31] . The reported accuracy is the mean of the accuracy metrics from ten experimental runs for each dataset. The accuracies of the various state-of-the-art methods reported in this paper are copied from HybridSN research paper[31] and Hybrid supplementary material. For instance, the accuracy figures reported in columns 2-7 on table 6, 7, and 8 are copied from hybrid supplementary material, while all rows information except the last one of table 11 is from the HybridSN research paper.

Finally, we generate the confusion Matrix for each dataset. The confusion matrix outputs a matrix that depicts the complete performance of the model. It illustrates the correctly and incorrectly classified samples at per class level. The accuracy of the matrix can be calculated by taking the sum of the values lying across the "main diagonal" divided by the total numbers of samples.

$$Accuracy = \frac{Correctly\ classified\ samples}{Total\ Number\ of\ samples} \quad (8)$$

Where, Correctly classified samples are cases where the predicted results are the same as the actual ground truth label. The Average Accuracy is given by:

$$Average\ Accuracy = \frac{1}{c}\sum_{i=1}^{c}(x) \quad (9)$$

Where c *is the number of classes*, and $x$ is the percentage of correctly classified pixels in a single class. Finally, the Kappa coefficient is as follows:

$$Kappa\ Coefficient = \frac{P_o - P_e}{1 - P_e} \quad (10)$$

Where, $P_o$ denotes the Observed agreement which is the model classification accuracy(see equation (8)) and $P_e$ symbolizes the expected agreement between the model classification map and the ground truth map by chance probability. When the kappa value is 1, it indicates perfect agreement while 0 indicate agreement by chance.

*1) The IP Dataset*

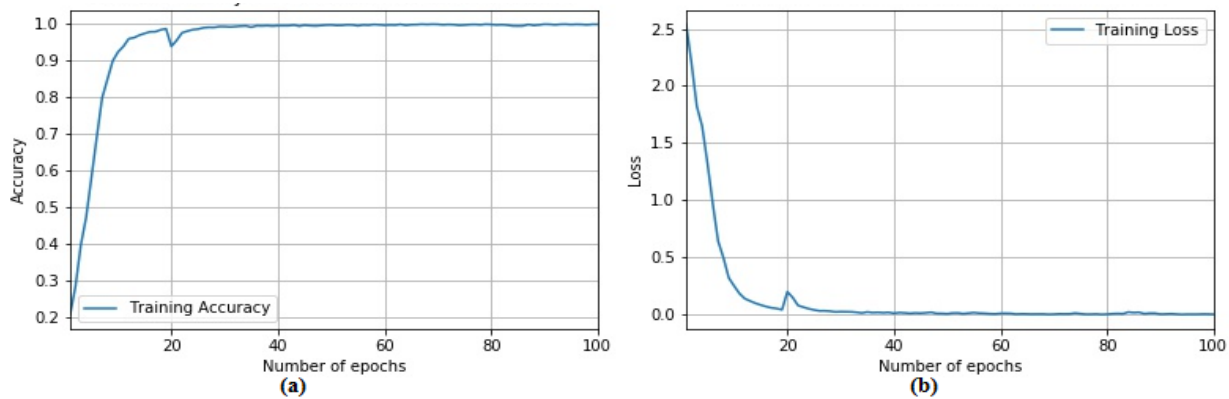

Fig. 4.(a) Model Accuracy with 30% train data on IP dataset, (b) Model Loss with 30% train data on IP dataset

Fig.4 shows our model training accuracy and loss graph for 100 epochs with 30% training data on the IP scene dataset. It can be observed that our model is computationally efficient as it quickly converges at approximately 50 epochs.

TABLE 3:
PER CLASS ACCURACY ON THE IP SCENE DATASET WITH 30% TRAIN SET.

| Class Name. | SVM | 2D-CNN | 3D-CNN | M3D-CNN | SSRN | HybridSN | **Ours** |
|---|---|---|---|---|---|---|---|
| Alfalfa | 82.2 | 75 | 79.23 | 97.03 | 97.82 | 99.38 | **100** |
| Corn-no | 73.82 | 81.4 | 88.6 | 97.9 | 99.17 | 99.58 | **99.66** |
| Corn-min | 82.15 | 87.6 | 85.81 | 92.41 | 99.53 | 99.66 | **99.91** |
| Corn | 77.12 | 62.04 | 90.53 | 93.25 | 97.79 | 99.88 | **100** |
| Grass-pasture | 73.66 | 92.3 | 96.11 | 95 | 99.24 | 99.53 | **99.82** |
| Grass-trees | 93.4 | 99.21 | 98.43 | 99.74 | 99.51 | **99.96** | 99.86 |
| Grass-pasture-mowed | 96.21 | 75 | 92.36 | **100** | 98.7 | 99 | **100** |
| Hay-windrowed | 85.72 | 100 | 98.51 | 99.99 | 99.85 | **100** | 100 |
| Oats | 97.38 | 64.28 | 88.9 | 96.61 | 98.5 | **100** | 95.71 |
| Soybean-no | 71.01 | 82.79 | 87.72 | 96.32 | 98.74 | **99.56** | 99.40 |
| Soybean-min | 76.5 | 91.27 | 91.42 | 97.13 | 99.3 | **99.84** | 99.78 |
| Soybean-clean | 83.9 | 82.89 | 90.04 | 97.16 | 98.43 | **99.52** | 99.18 |
| Wheat | 83.56 | 99.3 | 99 | 99.6 | **100** | 99.86 | 99.93 |
| Woods | 98.63 | 98.87 | 97.95 | 98.42 | 99.31 | **100** | 99.95 |
| Buildings-Grass-Trees-Drives | 94.21 | 86.29 | 82.57 | 83.31 | 99.2 | 99.85 | **99.96** |
| Stone-Steel-Towers | 69.63 | 100 | 98.51 | 100 | 97.82 | 98.46 | **99.54** |
| **OA** | 85.3±2.81 | 89.48±0.15 | 91.1±0.42 | 95.32±0.11 | 99.19±0.26 | 99.75±0.11 | 99.75±0.05 |
| **Kappa** | 83.1±3.15 | 87.96±0.51 | 89.98±0.5 | 94.7±0.2 | 99.07±0.3 | 99.71±0.13 | 99.72±0.06 |
| **AA** | 79.03±2.65 | 86.14±0.82 | 91.58+0.15 | 96.41+0.72 | 98.93+0.59 | 99.63+0.15 | 99.55±0.44 |

From table 6, it can be observed that our model performance at the class level on the Indian Pines dataset give the highest score in 9 out 16 classes. At the entire dataset level, our model yields a competitive result on Overall Accuracy (OA) and Kappa Coefficient (Kappa), with slightly lower Average Accuracy (AA) as compared to the state-of-the-art methods.

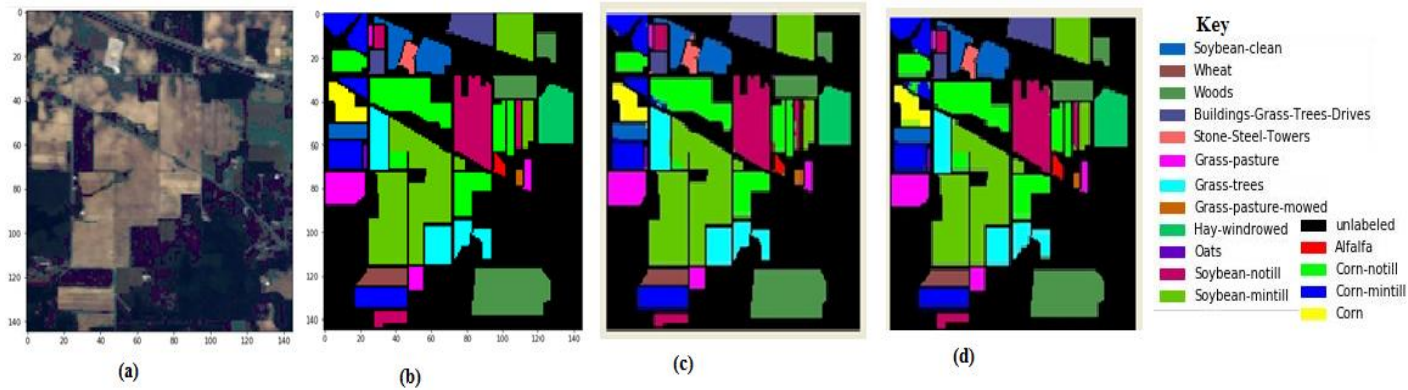

Fig. 5. (a) Original Image, (b) Ground Truth, (c) Prediction on 10% train data, (d) Prediction on 30% train data on IP dataset

Fig. 5 (a) contains the original image of the IP dataset. We can see that the ground truth (Fig,5(b)) of the IP dataset is comparable to the predicted images even with little training data (i.e. 30% of the total samples) to insufficient training data (i.e. 10% of the total samples).

2) *PU Dataset*

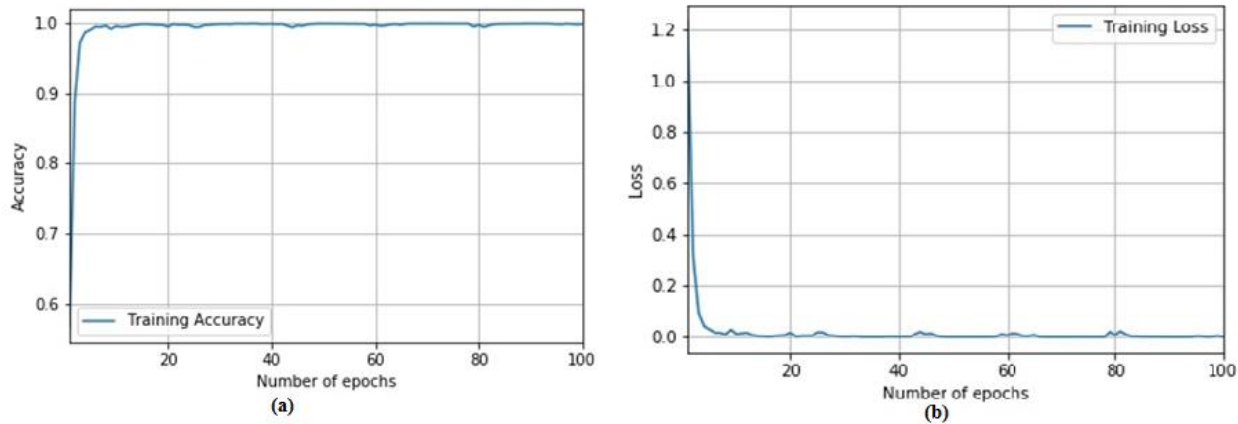

Fig. 6.(a)Model Accuracy with 30% train data on PU dataset, (b) Model Loss with 30% train data on PU dataset.

Fig.6 shows our model training accuracy and loss convergence graph for 100 epochs of 30% training data on the IP scene dataset. This proves that our model is computationally efficient as it drastically converges at approximately 20 epochs.

TABLE 4:
PER CLASS ACCURACY ON PU DATASET (30% TRAIN SET)

| Class Name | SVM | 2D-CNN | 3D-CNN | M3D-CNN | SSRN | HybridSN | Ours |
|---|---|---|---|---|---|---|---|
| Asphalt | 94.72 | 98.51 | 98.4 | 98.31 | **100** | **100** | **100** |
| Meadows | 97.15 | 99.54 | 96.91 | 96.1 | 99.87 | **100** | **100** |
| Gravel | 82.73 | 84.62 | 97.05 | 96.34 | 100 | **100** | 99.99 |
| Trees | 96.82 | 98.04 | 98.84 | 98.82 | **100** | 99.84 | 99.72 |
| Painted_metal_sheets | 99.71 | **100** | **100** | 99.97 | **100** | **100** | **100** |
| Bare_Soil | 90.48 | 97.1 | 99.32 | 99.83 | **100** | **100** | **100** |
| Bitumen | 87.73 | 95.05 | 98.92 | 99.66 | **100** | **100** | **100** |
| Self-Blocking_Bricks | 88.29 | 96.39 | 98.33 | 99.23 | 99.34 | 99.98 | **100** |
| Shadows | 99.9 | 99.69 | 99.9 | 99.92 | 100 | **99.9** | 99.88 |
| OA | 94.34±0.18 | 97.86±0.2 | 96.53±0.08 | 95.76±0.2 | 99.90±0 | **99.98+0.01** | 99.97+0.01 |
| Kappa | 92.5+0.7 | 97.16±0.51 | 95.51±0.21 | 94.5±0.15 | 99.87±0.0 | **99.98±0.01** | **99.98±0.01** |
| AA | 92.98±0.41 | 96.55±0.03 | 97.57±1.31 | 95.08±1.21 | 99.91±0.0 | **99.97±0.01** | 99.96±0.02 |

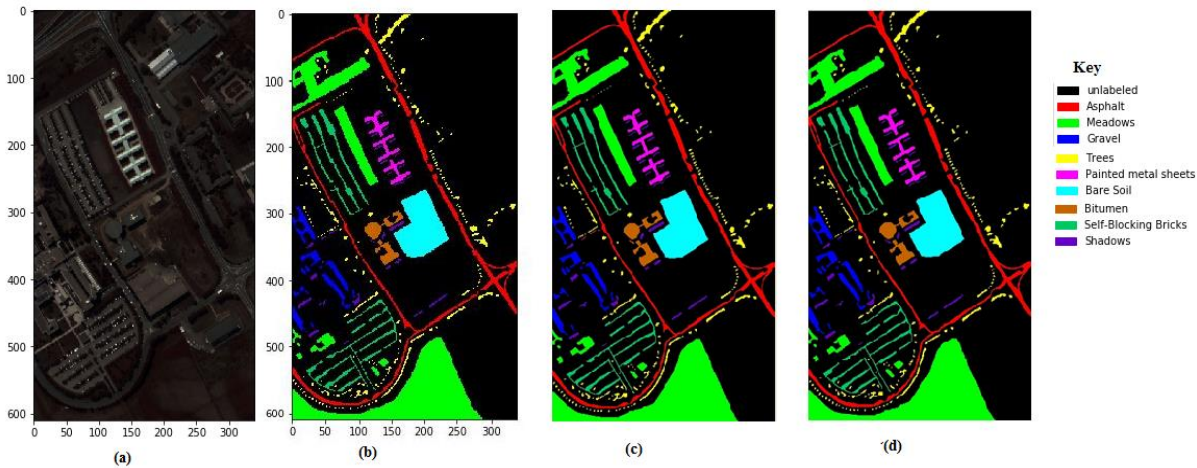

Fig. 7.(a) Original Image, (b) Ground Truth, (c) Prediction on 10% train data, (d) Prediction on 30% train data on PU dataset

Fig. 7 shows that the ground truth of the PU dataset (Fig7 (b)) is indistinguishable with the predicted maps generated using small (10% and 30%) training sample size. This shows that our architecture is robust in HSI image classification.

*3) The SA Dataset*

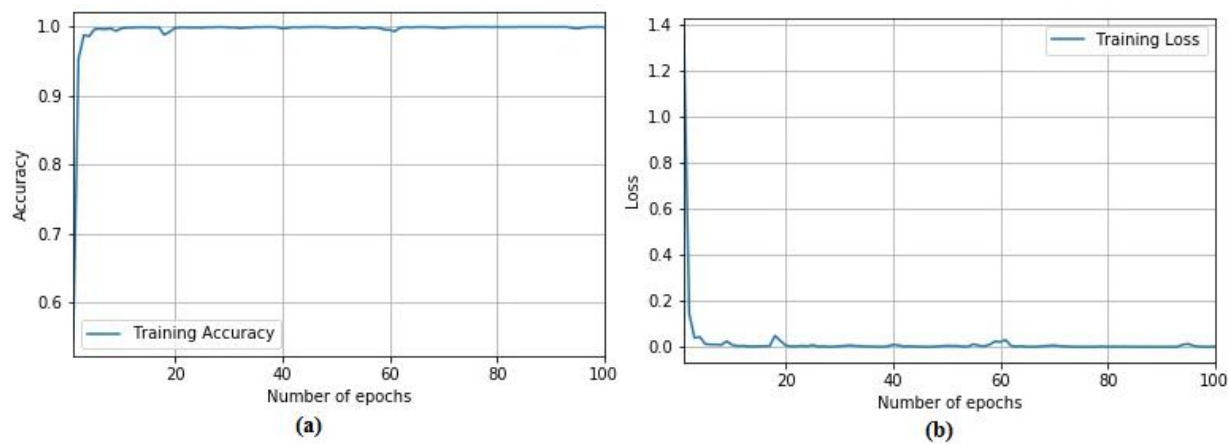

Fig. 8.(a) Model Accuracy with 30% train data on SA dataset, (b) Model Loss with 30% train data on SA dataset

Fig.8 illustrates our model training accuracy and loss graph for 100 epochs of 30% training data on the SA scene dataset. The model converges at around 30 epochs which shows that our model attains high computation efficiency.

TABLE 5:
PER CLASS ACCURACY ON SA DATASET (30% TRAIN SET)

| Class Name. | SVM | 2D-CNN | 3D-CNN | M3D-CNN | SSRN | HybridSN | Ours |
|---|---|---|---|---|---|---|---|
| Brocoli_green_weeds | 99.6 | 100 | 98.41 | 97.5 | 100 | 100 | 100 |
| Brocoli_green_weeds | 99.82 | 99.96 | 100 | 100 | 100 | 100 | 100 |
| Fallow | 99.26 | 99.63 | 99.23 | 99.43 | 100 | 100 | 100 |
| Fallow_rough_plow | 99.4 | 99.28 | 99.9 | 99.51 | 99.89 | 100 | 100 |
| Fallow_smooth | 99.42 | 99.2 | 99.43 | 99.72 | 100 | 100 | 100 |
| Stubble | 100 | 100 | 99.55 | 99.23 | 100 | 100 | 100 |
| Celery | 99.83 | 100 | 99.72 | 99.45 | 100 | 100 | 100 |
| Grapes_untrained_ | 85.25 | 93.62 | 89.75 | 92.63 | 100 | 100 | 100 |
| Soil_vinyard_develop | 99.71 | 100 | 99.81 | 99.7 | 100 | 100 | 100 |
| Corn_senesced_green_weeds | 97.03 | 98.82 | 98.36 | 97.31 | 99.91 | 100 | 100 |
| Lettuce_romaine_4wk | 98.24 | 99.73 | 98.12 | 98.05 | 100 | 100 | 100 |

| Lettuce_romaine_5wk | 99.46 | 100 | 98.96 | 98.5 | 100 | 100 | 100 |
|---|---|---|---|---|---|---|---|
| Lettuce_romaine_6wk | 98.77 | 100 | 98.93 | 98.7 | 100 | 100 | 100 |
| Lettuce_romaine_7wk | 97.3 | 99.86 | 98.6 | 98.42 | 100 | 100 | 100 |
| Vinyard_untrained | 72.71 | 91.52 | 79.31 | 87.18 | 99.96 | 100 | 100 |
| Vinyard_vertical_trellis | 99.41 | 99.92 | 94.51 | 91.11 | 100 | 100 | 100 |
| OA | 92.95±0.34 | 97.38±0.02 | 93.96±0.15 | 94.79+0.3 | 99.98+0.1 | 100±0.0 | 100±0.0 |
| Kappa | 92.11±0.18 | 97.08±0.1 | 93.32±0.5 | 94.2±0.22 | 99.97±0.1 | 100±00 | 100±00 |
| AA | 94.6±2.28 | 98.84±0.06 | 97.01+0.63 | 96.25±0.56 | 99.97±0.0 | 100±0.0 | 100±0.0 |

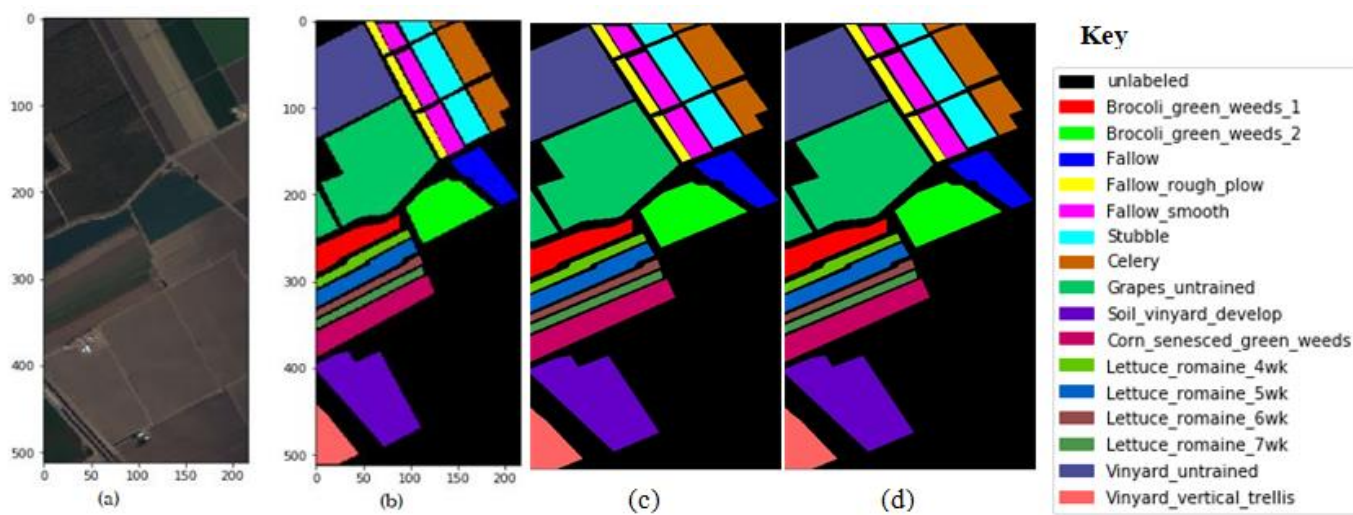

Fig. 9.(a) Original Image, (b) Ground Truth, (c) Prediction on 10% train data, (d) Prediction on 30% train data on SA dataset

*4) BW Dataset*

We can see from Fig. 9 that the Salinas scene dataset's predicted images using both 10% and 30% ((Fig. 9 (c, d)) training sample size generates identical classification maps to the provided ground truth (Fig.9 (b)). This demonstrates that our model has high classification accuracy on the Salinas Scene dataset.

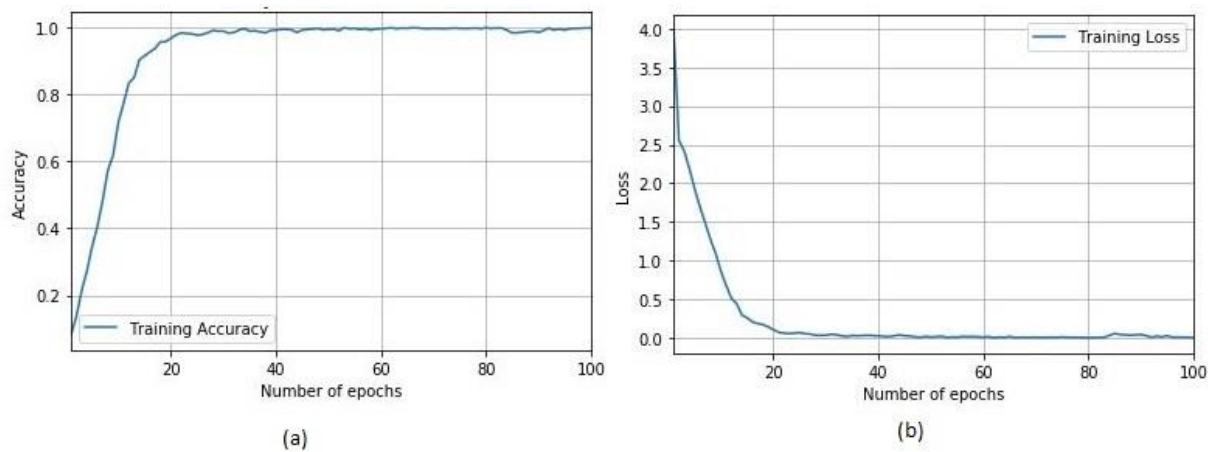

Fig. 10.( a) Model Accuracy with 30% train data on BW dataset, (b) Model Loss with 30% train data on BW dataset

Fig. 10 shows our model training accuracy and loss graph for 100 epochs with 30% training data on the BW dataset. The model attains convergence at around 50 epochs which confirms the fast learning of the model.

On table 10, we illustrate per class classification accuracy for the Botswana dataset. It's nontrivial to note that many Research work from the literature has not reported the accuracies on Botswana datasets. To show our model's performance on the Botswana dataset in comparison with the state-the-art methods, we use the result reported by Zhang et al [37]. The result clearly demonstrates our model better performance on the Botswana dataset that is characterized by low spatial resolution (30m). The model performance on the Botswana dataset, calls for more investigation on the application of HSI models on low spatial resolution multispectral satellite images. If findings turn positive, then the need for using multi-stream network architectures will be eliminated in satellite image classification tasks[38].

TABLE 6:
PER CLASS ACCURACY ON BW DATASET WITH 30% TRAIN SET

| Class No | Class Label | Method | |
|---|---|---|---|
| | | MSDN[37] | Ours |
| 1 | Water | 97.35 | **99.84** |
| 2 | Hippo grass | 100 | 100 |
| 3 | Floodplain Grasses 1 | 99.45 | **100** |
| 4 | Floodplain Grasses 2 | 100 | 100 |
| 5 | Reeds 1 | 96.76 | **99.31** |
| 6 | Riparian | 97.87 | **98.94** |
| 7 | Firescar 2 | 100 | 100 |
| 8 | Island interior | **100** | 99.72 |
| 9 | Acacia woodlands | 99.54 | **100** |
| 10 | Acacia shrublands | 100 | 100 |
| 11 | Acacia grasslands | 100 | 100 |
| 12 | Short mopane | **100** | 99.84 |
| 13 | Mixed mopane | 99.46 | **99.84** |
| 14 | Exposed soils | 100 | 100 |
| | OA | - | 99.79 ±0.2 |
| | Kappa | - | 99.80 ±0.2 |
| | AA | 99.32 | 99.82 ±0.2 |

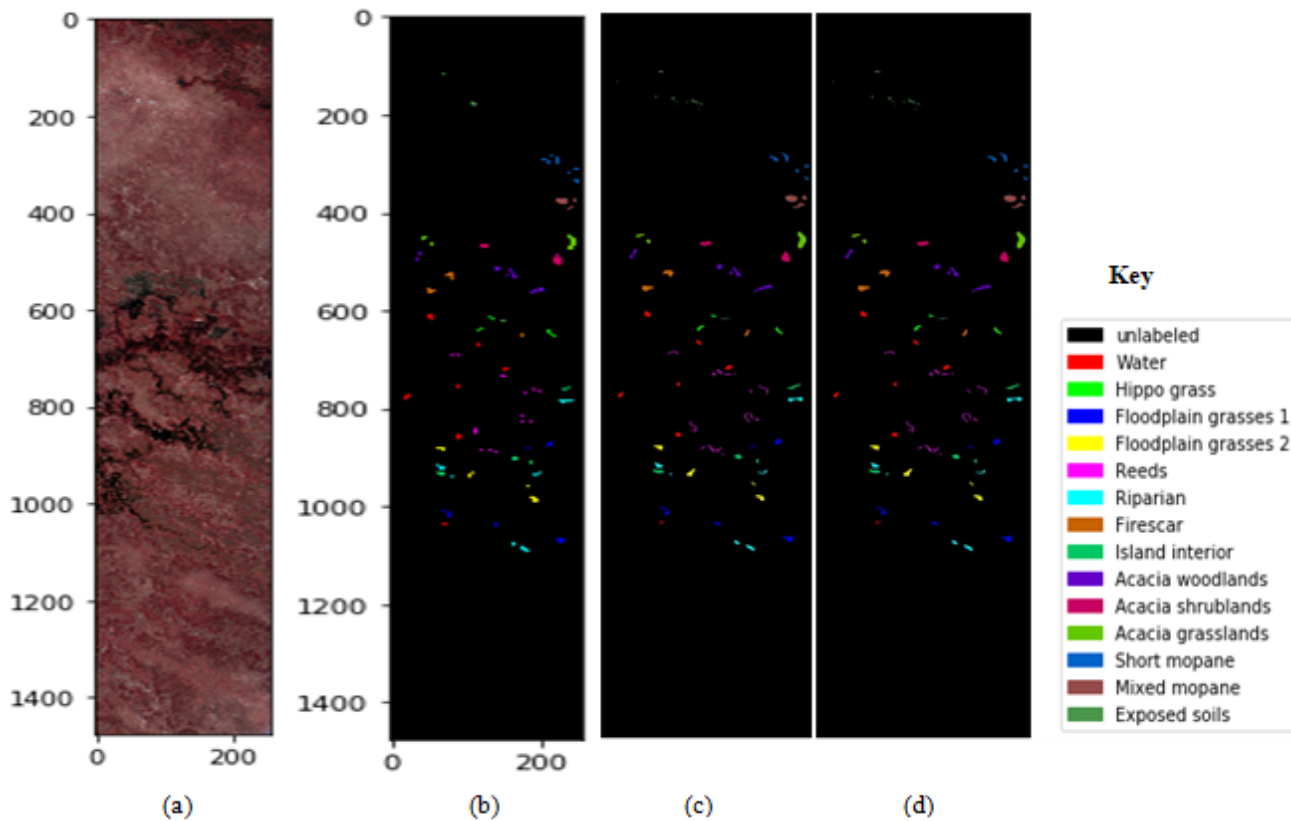

Fig. 11. (a) Original Image, (b) Ground Truth, (c) Prediction on 10% train data, (d) Prediction on 30% train data on Botswana dataset

Fig. 11 clearly illustrates that even with little training data (i.e. 30% of the total samples) to meager training data (i.e. 10% of the total samples) on the Botswana dataset; the predicted images are indistinguishable from the ground truth image. It is a clear indication that our model generates better-quality classification maps.

D. *Entire Dataset Accuracy*

TABLE 7:
THE CLASSIFICATION ACCURACIES (IN PERCENTAGES) ON IP, PU, SA USING PROPOSED AND STATE-OF-THE-ART METHODS WITH 30% TRAINING SAMPLE SIZE.

| Methods | IP | | | PU | | | SA | | |
|---|---|---|---|---|---|---|---|---|---|
| | OA | Kappa | AA | OA | Kappa | AA | OA | Kappa | AA |
| SVM | 85.30 ±2.8 | 83.10±3.2 | 79.03±2.7 | 94.34±0.2 | 92.50±0.7 | 92.98±0.4 | 92.95±0.3 | 92.11±0.2 | 94.60±2.3 |
| 2D-CNN | 89.48±0.2 | 87.96±0.5 | 86.14±0.8 | 97.86±0.2 | 97.16±0.5 | 96.55±0.0 | 97.38±0.0 | 97.08±0.1 | 98.84±0.1 |
| 3D-CNN | 91.10±0.4 | 89.98±0.5 | 91.58±0.2 | 96.51±0.2 | 95.51±0.2 | 97.57±1.3 | 93.96±0.2 | 93.32±0.5 | 97.01±0.6 |
| M3D-CNN | 95.32±0.1 | 99.07±0.3 | 98.93±0.6 | 95.76±0.2 | 94.50±0.2 | 95.08±1.2 | 94.79±0.3 | 94.20±0.2 | 96.25±0.6 |
| SSRN | 99.19±0.3 | 99.07±0.3 | 98.93±0.6 | 99.90±0.0 | 99.87±0.0 | 99.91±0.0 | 99.98±0.1 | 99.97±0.1 | 99.97±0.0 |
| HybridSN | **99.75 ±0.1** | 99.71±0.1 | **99.63 ±0.2** | 99.98±0.0 | 99.98±0.0 | 99.97±0.0 | **100 ±0.0** | **100 ±0.0** | **100 ±0.0** |
| **OURS** | **99.75± 0.1** | **99.72±0.1** | 99.55 ±0.4 | **99.98+0.0** | 99.97±0.0 | 99.96±0.0 | **100 ±0.0** | **100 ±0.0** | **100 ±0.0** |

Table 10 shows the entire dataset result summary with 30% training sample sizes on Indian Pines, Pavia University and Salinas Scene datasets. We use the Kappa coefficient, OA, and AA accuracy metrics to compare the performance of our model in relation to the state-of-art approaches. It is evident that our method achieves competitive accuracy with the HybridSN model accuracy across the three datasets (IP, PU, and SA). Moreover, our model maintains a minimum standard deviation across all the experimental datasets demonstrating its stability.

TABLE 8:
COMPARING OUR CLASSIFICATION ACCURACIES (IN PERCENTAGES) WITH VARIOUS STATE-OF-ART METHODS ON IP, PU, SA WITH 10% TRAINING SAMPLE SIZE

| Methods | IP | | | PU | | | SA | | |
|---|---|---|---|---|---|---|---|---|---|
| | OA | Kappa | AA | OA | Kappa | AA | OA | Kappa | AA |
| 2D-CNN | 80.27 ±1.2 | 78.26±2.1 | 68.32 ±4.1 | 96.63 ±0.2 | 95.53 ±1 | 94.84 ±1.4 | 96.34 ±0.3 | 95.93 ±0.9 | 94.36 ±0.5 |
| 3D-CNN | 82.62 ±0.1 | 79.25 ±0.3 | 76.51 ±0.1 | 96.34 ±0.2 | 94.9 ±1.2 | 97.03 ±0.6 | 85 ±0.1 | 83.2 ±0.7 | 89.63 ±0.2 |
| M3D-CNN | 81.39 ±2.6 | 81.2 ±2 | 75.22 ±0.7 | 95.95 ±0.6 | 93.4 ±0.4 | 97.52 ±1 | 94.2 ±0.8 | 93.61 ±0.3 | 96.66 ±0.5 |
| SSRN | **98.45 ±0.2** | **98.23 ±0.3** | 86.19 ±1.3 | 99.62 ±0 | 99.5 ±0 | **99.49 ±0** | 99.64 ±0 | 99.6 ±0 | 99.76 ±0 |
| HybridSN[31] | 98.39 ±0.4 | 98.16 ±0.5 | **98.01 ±0.5** | 99.72 ±0.1 | 99.64 ±0.2 | 99.2 ±0.2 | **99.98 ±0** | **99.98 ±0** | **99.98 ±0** |
| **Ours** | 98.44 ±0.2 | 98.22 ±0.2 | 97.91 ±0.7 | **99.73 ±0** | **99.65 ±0** | 99.24 ±0.1 | **99.98 ±0** | **99.98 ±0** | 99.97 ±0 |

Table 11 shows the results when comparing our architecture accuracies with various state-of-art approaches on IP, PU, and SA using very little training data (i.e. 10% of the total samples). It can be observed that our architecture performance is slightly superior to the state-of-art approaches in almost all cases while maintaining minimum standard deviation.

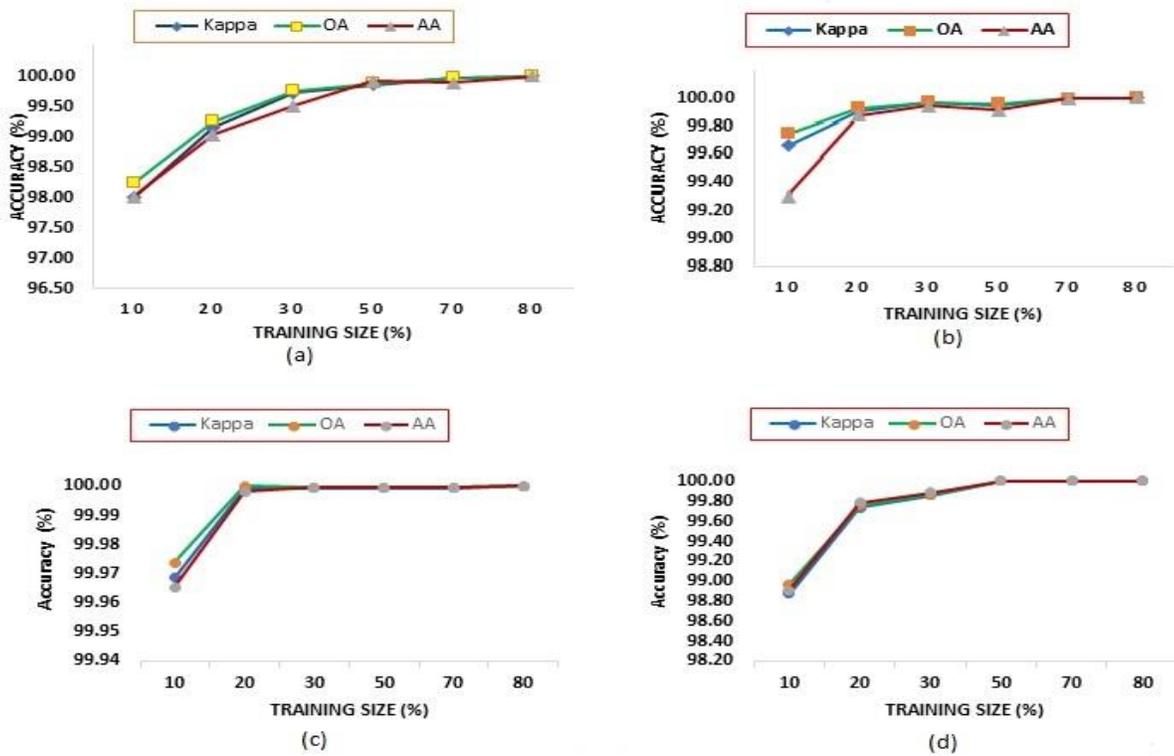

Fig. 12.The Kappa coefficient (Kappa), Overall Average (OA), and Average Accuracy (AA) for classification results on (a) Indian Pine, (b) Pavia University (c) Salinas Scene, (d) Botswana dataset against different training sample sizes.

Fig.12 (a) shows the OA, AA, and Kappa Coefficient of IP dataset increase exponentially when the training sample size is set to 10% and begins to plateau when the training sample size is 30%. A similar trend is observed in fig.12 (b) for the PU dataset. Fig 12 (c) shows a sharp increase in OA, AA, and Kappa Coefficient of SA dataset when the training sample size is set to 10% and begins to plateau when the training sample size is 20%. In Fig 12 (d) the OA, AA, and Kappa Coefficient of BS dataset increases drastically when the training sample size is set to 10% and plateaus when the training sample size is 50%. From the figure above, it shows that the test accuracy increases fast with the increase of training sample size 10% to 30%. From 30% the accuracy increase in minimal and the graphs flattens at 50%. Our optimal results are attained at a training-testing ratio of 3:7; this explains why we chose to report our accuracy at 30% of the training sample size.

Fig. 13. The Confussion Matrix for Indian Pine(a),Pavia University(b), Salinas Scene(c), Botswana(d) dataset

Fig. 13 shows the confusion matrix that depicts the complete performance of the model with 30% training sample data on IP (fig. 13(a)), UP (fig. 13(b)), SA(fig. 13(c)), and BW (fig. 13(d)), datasets, respectively. It can be observed that very high diagonal values (with bolded font) lie cross the "main diagonal" of the entire confusion matrices. This is an indication that our model is able to predict correctly at the class level with just a few incorrect outcomes (values with the yellow background color).

TABLE 9:
PARAMETER/ACCURACY COMPARISON ON THE IP DATASET

| No. | Model | Parameters | Accuracy Metrics | | |
|---|---|---|---|---|---|
| | | | OA | Kappa | AA |
| 1 | SVM | - | 85.30 ±2.8 | 83.10 ±3.2 | 79.03 ±2.7 |
| 2 | 2D-CNN | 966,346 | 89.48 ±0.2 | 87.96 ±0.5 | 86.14 ±0.8 |
| 3 | 3D-CNN | 46,107 | 91.10 ±0.4 | 89.98 ±0.5 | 91.58 ±0.2 |
| 4 | M3D-CNN | 284,897 | 95.32 ±0.1 | 99.07 ±0.3 | 98.93 ±0.6 |
| 5 | SSRN | 346,784 | 99.19 ±0.3 | 99.07 ±0.3 | 98.93 ±0.6 |
| 6 | HybridSN | 5,122,176 | 99.75 ±0.1 | 99.71 ±0.1 | 99.63 ±0.2 |
| 7 | **OURS** | **332,864** | **99.75 ±0.1** | **99.72 ±0.1** | **99.55 ±0.4** |

Table12 shows our model parameters compared to those of the state-of-the-art method. It's evident that our model achieves high accuracy with fewer parameters compared to the state-of-the-art approaches.

TABLE 10:
TRAIN AND TEST TIME COMPLEXITY ON IP AND PAVIA UNIVERSITY SCENE DATASETS GIVEN IN GPU TIME

| No. | Model | IP | | PU | |
|---|---|---|---|---|---|
| | | Train | Test | Train | Test |
| 1 | SSRN | 902.58 | 3.19 | 1837.72 | 10.27 |
| 2 | OURS | 459.24 | 2.60 | 585.49 | 7.22 |

Table 13 illustrates our method and the SSRN model time complexity. The result shows that our model has less computational time on both IP and PU datasets compared to the SSRN model.

V. CONCLUSION

This paper proposes the MixedSN model that extends the HybridSN and SSRN methods for hyperspectral image classification. The proposed MixedSN model introduces bottleneck layers that drastically reduce the number of parameters and show general implementation structure using ResNext network for HSI Deep learning models. Our proposed model is computationally efficient compared to the state-of-the-art methods. It also shows superior performance for small to insufficient training data. The experiments over four benchmark datasets compared with the recent state-of-the-art models confirm the superiority of the proposed method.

Moreover, we propose more research work on the application of HSI deep learning models on low-resolution multi-spectral satellite image classification. Our idea is based on our model performance on Botswana Dataset that is characterized by low spatial resolution.

The successful application of deep learning approaches to HSI data depends on the availability of large training samples. This experiment still suffers from the limited number of training samples available among the experimented datasets hence causes Overfitting.

arXiv : 7 February,2020    15[  ] classification," in *2nd Workshop on Hyperspectral Image and Signal Processing: Evolution in Remote Sensing, WHISPERS 2010 - Workshop Program*, 2010.

[3] T. V. Bandos, L. Bruzzone, and G. Camps-Valls, "Classification of hyperspectral images with regularized linear discriminant analysis," *IEEE Trans. Geosci. Remote Sens.*, vol. 47, no. 3, pp. 862–873, Mar. 2009.

[4] L. M. Bruce, C. H. Koger, and J. Li, "Dimensionality reduction of hyperspectral data using discrete wavelet transform feature extraction," *IEEE Trans. Geosci. Remote Sens.*, vol. 40, no. 10, pp. 2331–2338, Oct. 2002.

[5] L. O. Jimenez, "Hyperspectral data analysis and supervised feature reduction via projection pursuit," *IEEE Trans. Geosci. Remote Sens.*, vol. 37, no. 6, pp. 2653–2667, Nov. 1999.

[6] Y. Bengio, A. Courville, and P. Vincent, "Representation learning: A review and new perspectives," *IEEE Trans. Pattern Anal. Mach. Intell.*, vol. 35, no. 8, pp. 1798–1828, 2013.

[7] D. Lunga, S. Prasad, M. M. Crawford, and O. Ersoy, "Manifold-learning-based feature extraction for classification of hyperspectral data: A review of advances in manifold learning," *IEEE Signal Process. Mag.*, vol. 31, no. 1, pp. 55–66, 2014.

[8] T. Han and D. G. Goodenough, "Investigation of nonlinearity in hyperspectral remotely sensed imagery - A nonlinear time series analysis approach," in *International Geoscience and Remote Sensing Symposium (IGARSS)*, 2007, pp. 1556–1560.

[9] J. B. Tenenbaum, V. De Silva, and J. C. Langford, "A global geometric framework for nonlinear dimensionality reduction," *Science (80-. ).*, vol. 290, no. 5500, pp. 2319–2323, Dec. 2000.

[10] C. M. Bachmann, T. L. Ainsworth, and R. A. Fusina, "Improved manifold coordinate representations of large-scale hyperspectral scenes," *IEEE Trans. Geosci. Remote Sens.*, vol. 44, no. 10, pp. 2786–2803, 2006.

[11] B. C. Kuo, C. H. Li, and J. M. Yang, "Kernel nonparametric weighted feature extraction for hyperspectral image classification," *IEEE Trans. Geosci. Remote Sens.*, vol. 47, no. 4, pp. 1139–1155, Apr. 2009.

[12] P. Hu, X. Liu, Y. Cai, and Z. Cai, "Band Selection of Hyperspectral Images Using Multiobjective Optimization-Based Sparse Self-Representation," *IEEE Geosci. Remote Sens. Lett.*, vol. 16, no. 3, pp. 452–456, Mar. 2019.

[13] T. Dundar and T. Ince, "Sparse Representation-Based Hyperspectral Image Classification Using Multiscale Superpixels and Guided Filter," *IEEE Geosci. Remote Sens. Lett.*, vol. 16, no. 2, pp. 246–250, Feb. 2019.

[14] Q. Gao, S. Lim, and X. Jia, "Hyperspectral image classification using joint sparse model and discontinuity preserving relaxation," *IEEE Geosci. Remote Sens. Lett.*, vol. 15, no. 1, pp. 78–82, Jan. 2018.

[15] B. Tu, X. Zhang, X. Kang, G. Zhang, J. Wang, and J. Wu, "Hyperspectral Image Classification via Fusing Correlation Coefficient and Joint Sparse Representation," *IEEE Geosci. Remote Sens. Lett.*, vol. 15, no. 3, pp. 340–344, Mar. 2018.

[16] P. Gao, J. Wang, H. Zhang, and Z. Li, "Boltzmann entropy-based unsupervised band selection for hyperspectral image classification," *IEEE Geosci. Remote Sens. Lett.*, vol. 16, no. 3, pp. 462–466, Mar. 2019.

[17] N. Kruger *et al.*, "Deep hierarchies in the primate visual cortex: What can we learn for computer vision?," *IEEE Trans. Pattern Anal. Mach. Intell.*, vol. 35, no. 8, pp. 1847–1871, 2013.

[18] Y. Chen, H. Jiang, C. Li, X. Jia, and P. Ghamisi, "Deep Feature Extraction and Classification of Hyperspectral Images Based on Convolutional Neural Networks," *IEEE Trans. Geosci. Remote Sens.*, vol. 54, no. 10, pp. 6232–6251, Oct. 2016.

[19] G. E. Hinton and R. R. Salakhutdinov, "Reducing the dimensionality of data with neural networks," *Science (80-. ).*, vol. 313, no. 5786, pp. 504–507, Jul. 2006.

[20] A. Krizhevsky, I. Sutskever, and G. E. Hinton, "ImageNet Classification with Deep Convolutional Neural Networks," in *Advances in Neural Information Processing Systems*, pp. 1097–1105.